# Scientific Collaborations: Principles of WikiBridge Design

Éric Leclercq and Marinette Savonnet

University of Burgundy
Le2I Laboratory - UMR 5158
B.P. 47 870, 21078 Dijon Cedex - France
`Firstname.Lastname@u-bourgogne.fr`

**Abstract.** Semantic wikis, wikis enhanced with Semantic Web technologies, are appropriate systems for community-authored knowledge models. They are particularly suitable for scientific collaboration. This paper details the design principles of WikiBridge, a semantic wiki.

## 1 Introduction

Collaborative environments for scientific knowledge are essential for scientists to formalize their ideas and to develop theories collaboratively. Large numerous corpuses created in a collaborative effort have emerged in recent years. For example the encyclopedia Wikipedia contains human-readable articles with texts, images which are grouped into categories and access with a full text search engine. Despite the power of wiki, it is difficult to answer a specific query because of the purely textual information stored. As a result, automated tools cannot understand knowledge and thus exploit it. Semantic wikis offer an open environment which allows to aggregate knowledge produced by different teams involved in different research fields. It also allows interoperability among corpuses in case of annotations defined using ontologies. Nevertheless, representing and managing life sciences data and related knowledge require a deep understanding of life sciences concepts. Chen and Carlis in [2] have identified four challenging characteristics that make difficult modeling in life sciences : 1) complexity of data (heterogeneous, incomplete, uncertain, inconsistent, multi-dimensional); 2) domain knowledge barrier; 3) evolving knowledge and 4) modeling skills of actors.

Our main objective is to build a platform that allows: 1) to capture both free and controlled data; 2) to verify that annotations have meaning in relation to domain knowledge; 3) and to ensure a reasonable cost of developing and maintaining.

The rest of the paper is organized as follow: section 2 describes the state of art of semantic wikis in Life Sciences, section 3 presents the design principles of WikiBridge. Finally, section 4 concludes the paper.

## 2 State of art

Wikis are simple to use, flexible and do not impose a predefined workflow, but with growing amount of content in wikis, it becomes increasingly difficult to find relevant information. In traditional wiki, semantics is not explicit, but is implicitly described by links between articles and by the context of the link (surrounding text): wiki engine does not interpret the knowledge contained in links graphs. A Semantic Wiki is a wiki enhanced with Semantic Web technologies that made it an appropriate system for community-authored knowledge models by adding machine readable semantics. Moreover, a semantic wiki supports incremental formalization of knowledge and offers enhanced navigation and query capabilities. We get insight below into semantic wikis for life sciences.

A wiki page of wikipedia[1] maintains a list of bioinformatics wikis and several semantic wikis projects dedicated to Life Sciences. In Makna [3], users can create semantic content using RDF statements referencing pre-existing ontologies. A specific application, MannWiki, has been developed for sharing knowledge on micro-array in Life Science domain. BOWiki [5] could be considered as a gene function wiki that can be used to collaboratively edit biomedical knowledge, biological information about genes and the latest scientific results. OpenDrugWiki [6] uses a semantic wiki as a consolidation tool for integrating heterogeneous drug data available in different documents. LabService Wiki [4] is dedicated to the management of a production laboratory and adds workflow capabilities to the semantic wiki engine. Brede Wiki [7] contains information from published peer-reviewed neuroscience articles. Information in the template forms includes for instance ontologies for brain function (i.e. cognitive components), brain regions. Databases can be generated from templates by only extracting the logical structures. Moreover, Boulos in [1] presents some semantic wikis examples from health and healthcare sciences.

## 3 WikiBridge Design principles

WikiBridge is developed as a prototype tool to support knowledge engineers in collaboratively formalizing knowledge. In this purpose we have extended MediaWiki with some structural DBMS capabilities and semantic tools (see figure 1). WikiBridge's design principles are influenced by requirements given by Uren et al. [8].

WikiBridge offers an **interactive WYSIWYG editor** using AJAX technology to communicate with the server backend in addition to the traditional structured text editor (offered by MediaWiki) that supports interactive typing of links and resources. The structured data can be managed by Semantic Forms extension[2] for MediaWiki. **Documents can be in different formats** such as word files, graphic files, excel sheets, XML formats, OBO flat file.

---

[1] http://en.wikipedia.org/wiki/Portal:Gene_Wiki/Other_Wikis
[2] http://www.mediawiki.org/wiki/Extension:Semantic_Forms

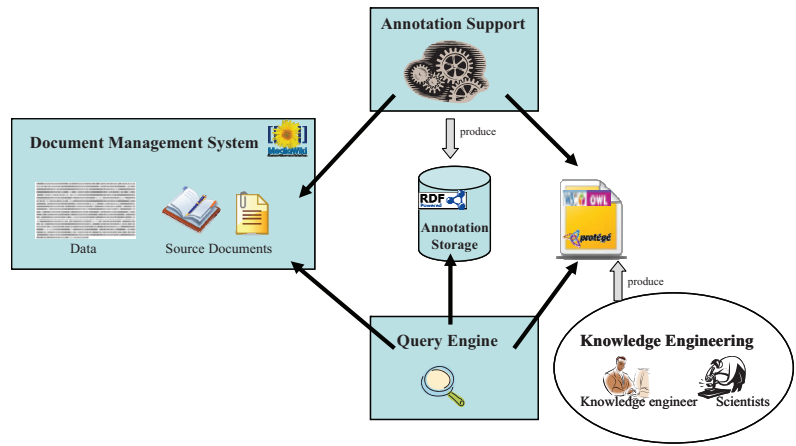

**Fig. 1.** An overview of WikiBridge

In a knowledge engineering process, it is common that non-technical domain experts work together with experienced knowledge engineers. This means that certain advanced functionalities can be hidden for novice users but are available to experienced users. Thus we use **Access Control List** (ACL) to describe privilege control depending on user identity and group affiliation.

WikiBridge supports **annotation functionalities within the same interface as the primary data** to avoid unnecessary context and application switches. In the meantime, these annotations must be maintained on clearly separated layers to keep integrity and traceability of the primary data. Different types of **annotations** have been identified: 1) simple annotation allows to tag a subject by describing some of its properties by a set of attribute value couples; 2) n-ary relation allows to map a subject with two or more values and references to other elements (subjects). In this case, some values properties reference another subject; 3) recursive annotation allows to explain or clarify an attribute by a sub-annotation. To be able to exchange data with other applications (e.g. ontology editors, Web Services, other wikis), WikiBridge is purely based on existing **Semantic Web standards** such as the Web Ontology Language OWL for describing ontologies and W3C's RDF for annotations. **Annotations are stored by RAP as RDF triples**, separately from the original document. We explicitly store in a database the conceptual model defining the structure of the domain ontology; ontology is loaded from Protégé by a specific program. **Information access** has been designed with taking into account some features about users. We have thus identified an usage typology in accordance to 1) kind of usage: reader, investigation, clarification; 2) knowledge degree of the domain: domain specialists like biologist researchers and non specialists. Queries are expressed in SPARQL. We consider **reasoning** as one of the most important functionalities as it allows : 1) to emerge knowledge that is not explicit in the data; 2) to check annotations, and 3) to enhance navigation and search. First order logic constraints are checked with a Java component that interacts with Pellet and Jena. WikiBridge extension connects to the constraint checker by the means of a web service.

## 4  Conclusion

In this article, we have presented WikiBridge, a feature-rich semantic wiki. The first experiment of the use of WikiBridge[3] shows many interesting possibilities for scientific community, mainly the possibility to scientists share and collaboratively build annotated knowledge. We have demonstrated that flexibility and data quality required by scientific applications can be achieved by using wiki with semantic web technologies.

---

[3] We have tested WikiBridge in 2009 and 2010 with a real project in archeology: Corpus Architecturae Religiosae Europeae - IV–X saec. ANR-07-CORP-011